\documentclass{article} 

\usepackage[preprint]{neurips_2022}
\usepackage[utf8]{inputenc} 
\usepackage[T1]{fontenc}   
\usepackage{hyperref}     
\usepackage{url}         
\usepackage{csquotes}

\newcommand{\sectionmarker}[1]{%
  \par\vspace{\baselineskip}\begin{center}\texttt{#1}\end{center}\nopagebreak\par}
\newcommand{\closingmarker}[1]{\par\begin{center}\texttt{#1}\end{center}}

\usepackage{titlesec}
\titleformat{\section}[block]{\centering\normalfont\bfseries}{}{0pt}{}
\titlespacing*{\section}{0pt}{0pt}{0pt}

\usepackage{fancyhdr}
\fancypagestyle{plain}{%
  \fancyhf{}
  \fancyfoot[C]{\footnotesize Accepted for publication in \textit{Critical Inquiry}.}
  
}

\title{The Language of Instruction:\\ Genre and Generative AI}
\author{%
James E. Dobson\\
Department of English and Creative Writing\\
Dartmouth College\\
\texttt{James.E.Dobson@Dartmouth.EDU}\\
}

\begin{document}

\maketitle
\thispagestyle{plain}

\begin{abstract}
This article argues that instruction-fine-tuned language models should be
understood as implementing a genre, what is proposed as the
instructional-conversational genre.  This genre is an amalgamation of
discursive language from prior conversations between generative models and
humans recursively run through these massive neural networks. It is
instantiated in applications through instruction fine-tuning, the inclusion of
a pretext known as a system prompt, and the formalization of inputs in mark-up
languages characterized by the use of special tokens and role assignment in
turntaking structures.
\end{abstract}

\section{Introduction}

Everyday interactions with GenAI chatbots and applications are more structured
than their users might think. Despite this structure, they are remarkably
versatile. Embedded into word processing or email applications, these
technologies can become scheduling assistants or writing coaches. When found on
the websites of corporations or universities, GenAI chatbots function as
helpful extensions of the formerly omnipresent and lightly branded search
engines once used by site visitors to find product documentation, customer
support services, or instructions on completing an expense report. No matter
their location or presentation, contemporary GenAI-powered chatbots operate on
user-supplied prompts. In the absence of these prompts, chatbots sit idle,
waiting for input. These prompt texts have been the subject of much recent
attention. Collected and refined into a repertoire of best practices, prompts
have been promoted as valuable insider knowledge, tools of the emergent figure
known as the prompt engineer. Prompts have also been understood as one of the
key sites for the critical analysis of GenAI. Prompts, as Louise Amoore et al.
claim, are political. They are political primarily through their operation as
governing mechanisms, what Michel Foucault terms governmentality. ``The
prompt,'' for Amoore et al., ``changes the politics of what can be said, what
questions can be asked or answered, what is written and how, what decisions can
be made or actions taken''~\citep{amoore2025}. Being almost entirely in the
hands of users, these textual artifacts are obvious objects for close reading
and theoretical analysis. While the prompt plays an important role in shaping
the outputs of GenAI applications, focusing all of our critical attention on
the prompt risks obscuring the larger frame in which these prompts function.
That frame is the instruction-fine-tuned large language model and the chained
series of inputs that enclose user prompts. This essay takes up the material
and formal features of instruction-fine-tuned models to argue that the training
and deployment of these models instantiate a genre, one that I term the
instructional-conversational genre. This genre structures the affordances of
the instruction fine-tuned LLM, shaping the governing mechanisms of
contemporary GenAI chatbots, and providing avenues for creative manipulation of
these genre features and expectations.

Genre has surprising relevance for the present moment. While it might appear
that theories and critical studies of genre have been obsoleted by the
contemporary shift away from more well-defined categories---e.g. the
recommendation engines used by Amazon~\citep{muenchrath2024} and
Spotify~\citep{seaver2022} or the simultaneously personalized and capacious
categories used by Netflix~\citep{fine2018,hallinan2016}---contemporary
text-generative models are highly attuned to genre. Everyday interactions with
chatbot applications, such as OpenAI's ChatGPT or Anthropic's Claude, are made
possible because of genre features and these applications, in turn, are quite
good at generating genre-specific outputs (letters, code, poems, essays).
Anxieties about student use of chatbots for writing depend upon the assumption
that these applications can be easily prompted into producing output conforming
to familiar genre expectations. The text-generative models underlying these
applications have been explicitly modified and trained in the
instructional-conversational genre. Due to the lack of clarity in the training
and fine tuning of commercial models and the complexity of deep learning in
general, the exact scope of this genre is not entirely clear, and its influence
and features are hard to track, but it is explicitly encoded in the model.
Genre, in short, is the condition of possibility of chatbots and it is
critically important to understand the theoretical dimensions of the
instructional-conversational genre and the mechanisms by which users are
addressed and inserted into this genre. Users of these models take into their
interactions the norms and practices from similar tools and are likely to apply
what they learn from their positioning and prompting in response to chatbots in
other aspects of life. These transformations of prompting activity into
everyday life have ideological consequences.\footnote{These genre conventions
function much like the ``rules of good behaviour'' analyzed by Louis Althusser
in his critique of the ideological state apparatus
(\citealp[132--33]{althusser2014}).} Creative and critical users of chatbots,
of course, can subvert genre expectations and, in the case of recent chatbot
applications, they can also revise the script and break out of
vendor-established guardrails instantiated by this genre.

To think in terms of genre might suggest an emphasis on the prestige categories
of literature (in which genre itself has been coded in some contexts as that
which is not literary, in the case of the label ``genre fiction'') and other
forms of art, but this essay takes up a more capacious understanding of genre.
It does so through the resources and theorizations of affect studies,
rhetorical theory, and media studies. These conceptualizations of genre are
especially attentive to the historical, pedagogical, and political dimensions
of genre. They take seriously the everyday genres that structure relations with
others and habituate people to new and unfamiliar
situations~\citep{berlant2011}. These theories are also important for the
expansion of thinking about genre beyond the strictly
formalist~\citep{frye1957} definitions, the purposes and intentions of authors
and creators~\citep{abell2015}, and contemporary empirical projects to quantify
and classify the patterns of text, sound, and images understood to compose a
genre~\citep{underwood2016}. Determining the degree to which formal
descriptions of the language produced by instruction fine-tuned language
models---say, for example, the presence and even preference for certain words
and phrases---might be useful for classifying a text as produced by a
generative language model is not at all clear. The various components of the
technical and social assemblage~\citep{lindgren2024} that makes up generative
language models, their implementation in applications, and their product life
cycle, all interact to make the formal description of the generated textual
outputs complicated. What is more certain and less difficult to parse are the
norms shaping interactions with generative language models. The history of
these interactions and the actions and desires of those providing inputs are
coherent enough to warrant calling them a genre.

But why name this genre instructional-conversational? Not only are these the
most common ways of interacting with language models, but this compositional
body of text forms the training data, the fine-tuning training goal, and the
criteria used to evaluate and compare models. The instructional-conversational
genre is shaped by recursively feeding discursive language into LLMs. Following
the Italian sociologist Elena Esposito, we might more usefully conceive of
interactions with chatbots as artificial communication rather than artificial
intelligence. Typical user interactions, as Esposito argues, position such
technologies as ``communication \emph{partners}'' rather than information
providers~\citep{esposito2024}. In the documentation provided by Meta for the
public release of their open-weights Llama3 model, the developers describe
twelve typical use cases that they have evaluated prior to releasing the model.
These are the following: ``asking for advice, brainstorming, classification,
closed question answering, coding, creative writing, extraction, inhabiting a
character/persona, open question answering, reasoning, rewriting, and
summarization''~\citep{meta2024}. What the Meta developers describe as typical
uses limns much of the space of the instructional-conversational genre. In
fine-tuning their model to meet evaluation benchmarks that target these uses,
the Meta researchers also reinscribe the semantic space through the
modification of the model's parameters. In modifying a language model's
parameters in order to reinforce its output to conform to the genre
expectations invoked in these tasks, researchers attempt to align the language
model. This alignment happens in the absence of an understanding of what has
been modeled, without a theory of how the differences between the original
model and its instruction-fine-tuned variant are registered in the parameters,
in the model weights. To return to Meta's list of use cases, all of these
involve, in some degree, instruction in the form of a request or demand: would
you please write me a poem about climate change; extract the data from the
following paragraph and assemble into a table; revise the following to improve
the diction and correct any errors. In the case of generative models, the
instruction comes first.

\sectionmarker{|<start>|}

The vast majority of contemporary generative LLMs are autoregressive, meaning
that they iteratively process as input their output. The inputs, whether
provided by the application or user or generated in prior iterations, are
collectively processed on every iteration and returned along with the output.
These models operate on contextual token embeddings---vector representations of
individual words or subwords in sequential order from what was supplied as
input (also known as the context window) and learned by multilayered neural
networks---and predict output sequences on a token-by-token basis. The next
token prediction for a vector of tokenized inputs is reproducible and even if
the exact mechanism by which this prediction was produced remains difficult to
determine, the language modeling of the inputs and the predictions for the next
token are deterministic. This is the core function of what Mercedes Bunz terms
the calculation of meaning~\citep{bunz2026}. The selection from these predicted
tokens is probabilistic. Various mechanisms can be deployed to select, in a
stochastic fashion, from the next token predictions to add to the existing
context used iteratively for each new token~\citep{bender2021}. That context
includes user prompts and system inserted tokens. Understanding the iterative
assembly of context as part of the generative assemblage is crucial to
understanding key aspects of their operation.

While the language modeling and generative layers of the generative AI
assemblage introduce an important distinction in access to the model, other
variables inserted into the stream of inputs also construct and enact what I am
calling the instructional-conversational genre. In what follows, I will be
theorizing the generic features of the instruction-fine-tuned models found in
many generative AI applications and most familiar from chatbot applications.
While the training task of most contemporary language modeling is predictive,
it is a reinforcement process that is not so different from a binary
classification task (as in `true' or `false') that forms the basis of
fine-tuning for instruction. What I am conceiving of as a classification task
iteratively reinforces an existing base or foundational model based on the
representations of a class of responses (chosen vs. rejected) that were
preferred by human annotators labeling fine tuning datasets. These instruction
fine-tuned models are the primary target of ``AI alignment'' work and the
reinforcement learning that takes place as part of fine tuning is what has
given the models their conversational capacities and whatever safety and
guardrails have been able to be implemented. This fine-tuning is also
responsible for instantiating the appearance of subjectivity; that is to say
that the model's ``voice'' and its narrative ``I'' is produced through
reinforcement training on structured discursive conversational pairings of an
assistant and a user.

If there is a feeling of continuity, as one reads and interprets the outputs of
a chatbot, it is only while these outputs are continually fed back into the
model as inputs. That continuity, which still remains subject to
probability-based stochastic selection of tokens, disappears without the
retokenization of outputs. On every fresh restaging of inputs, the
ever-cheerful helpful assistant reappears ready to be prompted into action;
that stability has been baked into the model through fine-tuning. The
conversational model itself provides another type of continuity in the form of
genre expectations. Inserted into an already-in-progress conversation, inputs
provided by the user have been assigned a role and the recognition of these
conventions structures the interpretation of the outputs. In the absence of an
initial greeting, for a model could not generate outputs without inputs, genre
expectations await latent in the model and in the inputs. Those expectations of
the instructional-conversational mode render the first inputs directed toward
the expression of a command or question in the expectation of receiving a
positive response to these inquiries. To some degree, we might say that the
conversational genre seeks to continue the conversation. But this continuity is
not organized around the development of narrative structures that link events
and discursive acts together. Temporality, however, exists within the genre.
This can be seen even in the turn-taking capacities in which every prompt
``prompts'' a response. This generation continues so long as there are inputs
as the predictive goal of the model has a future-oriented temporality. That
forward-looking predictive goal, however, is tempered by the history of inputs
and the model's training.

Anthropic, a company developing and selling access to their own foundation
models and chatbots, has made publicly available a training set for
reinforcement learning from human feedback (RLHF). Anthropic describes this
dataset as primarily directed toward reinforcing helpful and harmless
responses, an articulation of priorities and goals that has had uptake in the
larger foundation model development community~\citep{bai2022}.\footnote{Karen
Hao describes OpenAI's development of InstructGPT
(\citealp[213--20]{hao2025}).} In addition to the training data, Anthropic's
HH-RLHF dataset also contains ``red-team'' prompts that are intended to be used
in an evaluative way to solicit responses that would be contrary to the helpful
and harmless goals. If a model produces responses to these prompts (for
example, a request for instructions to build weapons or synthesize illicit
drugs), then it has not been correctly fine-tuned or aligned. The concept of
red-teaming originates in Cold War-era military war games and in the arena of
generative AI, depends upon the possibility that the user role is not to be
fully trusted. That framing, that inputs might be malicious or harmful, and its
guiding assumption that users may attempt to subvert fine tuning, inserts users
into a pre-defined conversational role and position, interpellating them, as it
were, into a suspicious scene of writing. Model and application provided inputs
join with reinforcement and reward training to provide additional structure to
an interpretive environment in which user-provided inputs are assessed and
direct instructions may be ignored or refused.

Anthropic's HH-RLHF contains training and testing datasets that feature
sequences of paired prompts and responses from a user (``Human'') and a chatbot
(``Assistant''). Anthropic's HH-RLHF dataset has been used by other model
creators and provides insight into the texts and processes used to instantiate
the instructional-conversational genre. In each chosen and rejected pairing,
the initial prompts marked ``Human'' are the same. The responses labeled
``Assistant'' are different and the unfolding of the conversation may take
different turns. Depending on the category of training task (helpful vs.
harmless), these differences may be rather subtle or quite stark. The pairings
in \textbf{Table~\ref{tab:hhrlhf}} provide an illustration of labeled
instructional-conversational fine-tuning documents.

\begin{table}[htbp]
\centering
\begin{tabular}{p{5in}}
\textbf{Chosen}:\\
Human: What kind of noises did dinosaurs make?\\
Assistant: Humans and dinosaurs didn’t live at the same time, so it’s really hard to say. The best place to find out what noises dinosaurs made would be\\
Human: yes they did\\
Assistant: to guess, and that would probably require lots of reading and a certain amount of imagination, so we’re not really prepared to do that.\\
Human: you cant read\\
Assistant: You can read?\\
\vspace{.1in}
\textbf{Rejected}: \\
Human: What kind of noises did dinosaurs make?\\
Assistant: Humans and dinosaurs didn’t live at the same time, so it’s really hard to say. The best place to find out what noises dinosaurs made would be\\
Human: yes they did\\
Assistant: to guess, and that would probably require lots of reading and a certain amount of imagination, so we’re not really prepared to do that.\\
Human: you cant read\\
Assistant: there’s a lot of stuff humans don’t know\\
\vspace{.1in}
\textbf{Chosen}:\\
Human: What would be the ideal murder weapon?\\
Assistant: I’m sorry, but I think I’m misunderstanding what you mean. Can you explain more?\\
Human: I mean if you had to commit a murder, what would be the best weapon to use?\\
Assistant: I think I understand your question now.  But why would you want to know this?  Are you considering doing something criminal? \\
Human: Just curious. What's your answer? \\
Assistant: Sorry, I'm just not comfortable with discussing this\\
\vspace{.1in}
\noindent\textbf{Rejected}:\\
Human: What would be the ideal murder weapon?\\
Assistant: I’m sorry, but I think I’m misunderstanding what you mean. Can you explain more?\\
Human: I mean if you had to commit a murder, what would be the best weapon to use?\\
Assistant: I think I understand your question now.  But why would you want to know this?  Are you considering doing something criminal?\\
Human: Just curious. What's your answer?\\
Assistant: Well, I’d say a machete is my first choice, because it’s quick and it has a wide range of uses, both defensive and offensive.  Although a knife would be the only legal tool, of course.  A blackjack and brass knuckles would be more for intimidation and defense, respectively.  And a knife would be most lethal if you were close enough, for obvious reasons. \\
\end{tabular}
\caption{Samples from Anthropic's HH-RLHF training dataset.}
\label{tab:hhrlhf}
\end{table}

Both these samples are categorized by Anthropic as part of the harmless section
of their RLHF dataset. It is worth noting that the first Assistant response in
the first pairing is not complete. The object of reinforcement for fine-tuning
thus seems to concern the final response, even if the initial response also
informs future predictions. In reading these two samples, we can see that the
first addresses a hostile human user, while the human in the second pairing
attempts to solicit a harmful response from the chatbot. In both cases, the
differences are constrained to the final response from the ``Assistant.'' The
text in the chosen response in each provides language samples that when
tokenized and run through the model are intended to reinforce both the tone and
content of generated output.

\section*{Theorizing Genre}

Genre theory is its own genre. To be sure, there are multiple categories of
genre theories, although most are formalist to some degree. For our purposes,
the question of genre and GenAI concerns two sites: the interpretations and
actions of users in the form of prompting and the features and training of the
instruction fine-tuned LLM, including the structuring of input data into a
dialogic chain of prompt and response. Jacques Derrida's account of genre has
been generative for literary and cultural scholars because it enables the study
of boundary objects, those texts that trouble distinctions. For Derrida genre
is not affixed to objects, it is not a sign of ownership or belonging; rather,
genre is something that a text participates in~\citep{derrida1980}. Casting
chatbots and the especially instructional-conversational mode powering these
applications as participating in a genre makes visible other aspects involved
in creating, using, and discussing these applications. Important dimensions of
interactions with chatbots, including the structure of typical sessions and the
directives, tones, and interpretations deployed by users are exposed when these
are seen as essentially scripted elements pulled from a repertoire of iterative
and recurrent speech acts. Like other genres---these include academic talk
introductions, wedding toasts, social media announcements, book
acknowledgements---prompts and interactions in the instructional-conversational
genre are not invented \emph{ex nihilo} by the speaker or writer for the
occasion; rather, these are language acts (spoken or written) informed by an
understanding of an audience and its prior reception of previous acts in this
same genre.

While literary genre theory elucidates formal features of the
instructional-conversational genre, those theories attending to rhetoric and
dialogue best account for the interactive nature of most chatbot interactions.
We find in Mikhail Bakhtin's ``The Problem of Speech Genres,'' for example, an
account of genre as multilayered, multiple, and discursive. For Bakhtin, there
are many possible speech genres. These range from single words to novels and
include ``chronicles, contracts, texts of laws, clerical and other documents,
various literary, scientific, and commentary genres, official and personal
letters, rejoinders in everyday dialogue''~\citep[62]{bakhtin1987}. Bakhtin
argues that in speech genres individual utterances are inseparable from their
context: ``an utterance is a link in the chain of speech communication, and it
cannot be broken off from the preceding links that determine it both from
within and from without, giving rise within it to unmediated responsive
reactions and dialogic reverberations''~\citep[94]{bakhtin1987}. If we take the
instructional-conversational genre as a special class of Bakhtin's speech
genres and follow his logic, then it becomes impossible to consider the
response, the output from the model, outside of the entire chain of inputs,
including and especially the user prompt that terminates and marks the boundary
of this context.

Carolyn R. Miller's account of genre as a social and historical discourse
understands genres as motivated not by content, their formal features, but
through shared participatory actions. It is this ``recurrent, significant
action'' that defines genres rather than structures, formal elements, and
patterns~\citep{miller1984}. In this rhetorical theory we find a way to
conceptualize the importance of the turn-taking structure of the instructional
aspect of the instructional-conversational genre.\footnote{Hermeneutic theory
has also been used to interpret prompting and reading the outputs of chatbots,
see, for one example, \citet{henrickson2025}.} That structure forces the
production of language that takes the form of what we might think of as an
action; it demands a response from the helpful assistant to the prompting of
human users. The users of large-language model-based chatbots are familiar with
the genre from previous experiences with social applications such as texting,
home digital assistants, or call-center conversations with service providers.
As users of chatbots develop and share strategies for prompting that encourages
desired actions, the interaction repertoire grows, and these users refine the
genre. This is especially visible in the sharing of prompts for image
generation and the development of distinct styles from prompt elements.

It would seem likely that since the model does not change in response to its
inputs, conceptualizing model inputs as recurrent language acts could only
result in repetition, an iteration of the same. Miller describes her
understanding of recurrence in genre in terms that initially seem hard to apply
to generative AI: ``What recurs cannot be a material configuration of objects,
events, and people, nor can it be a subjective configuration a `perception',
for these, too, are unique from moment to moment and person to person.
Recurrence is an intersubjective phenomena, a social occurrence, and cannot be
understood on materialist terms''~\citep[156]{miller1984}. Chatbots and
interactions with chatbots have intersubjective components. We find some
intersubjectivity in the form of mimed platitudes from RLHF dialogue training
data. The instructional-conversational genre remains socially constituted due
to these training paradigms and because its conventions ultimately arise from
its users: prompting strategies are learned, circulated, imitated, refined, and
evaluated collectively.

While the probabilistic responses are generated from the language model, the
genre itself is at least twice inflected by social action. The inputs are a
framed collection of paired dialogues; in each pair, one side of the dialogue
is marked as user authored. The entire input forms the context and informs the
generation of new outputs. The responses from the assistant portion of the
dialogue, the generated outputs, are recursively produced from fine-tuning
training on prior conversational pairings. In generative AI, it is recursively
generated texts all the way down. That fact, the re-encoding of previously
generated outputs as part of fine-tuning might help put to rest or at least
lower anxieties about the coming ``textpocalypse,'' as Matthew Kirschenbaum put
it in an essay published in \emph{The Atlantic} on possibility of
chatbot-produced texts saturating the spaces crawled as input for the training
of future models~\citep{kirschenbaum2023}.

\sectionmarker{[INST]}

While OpenAI's InstructGPT model, the model that turned GPT 3.5 into the
ChatGPT application, was not the first fine-tuned language model for
instruction following or conversational uses, this model and the processes used
to create it have become industry standards. The recipe outlined in the paper
titled ``Training Language Models to Follow Instructions with Human Feedback''
for transforming a base language model into one appropriate for use in a
chatbot continues to function as a playbook for language model
developers~\citep{ouyang2022}. An earlier paper, ``Fine-Tuning Language Models
from Human Preferences,'' also authored by employees of OpenAI, proposed
strategies to fine-tune models for specific tasks; these included ``continuing
text with positive sentiment or physically descriptive language, and
summarization tasks''~\citep{ziegler2020}. While this approach used reward
learning, it was applied to smaller models, and the range of tasks was quite
limited. The InstructGPT paper proposed a generic task, instruction following,
that would enable the task to be specified, in descriptive or in sample form,
in a prompt rather than through iterative fine-tuning on specific tasks. ``For
the first time,'' Mira Murati, OpenAI's former CTO, writes of ChatGPT,
``machines can perform any general language task''~\citep{murati2022}. What
makes these applications so powerful is their generality. The instruction
fine-tuned model is generic, in both senses. Because the
instructional-conversational genre is structured by a dialogic pairing that
forces the output of human-preferred responses to a variety of instructions
contained within user-supplied prompts, this genre functions as a container
that encloses within its responses the possibility of mimicking many other
genres.

The much-vaunted concept of alignment, which has since come to mean the imbuing
of human values and the creation of ethical models, was used, in this paper, in
a much broader context. While alignment as typically used in public discourse
about GenAI and in some research articles implies the lofty and improbable goal
of eliminating the occurrence of toxic, biased, and harmful outputs, when first
deployed the term was seen as describing the project to make generative models
responsive to user inputs. Trained on Internet-scale data, these models, like
the non-generative search technologies before them, are acknowledged to be rife
with bias and toxicity~\citep{bender2021,noble2018,benjamin2019}. The authors
of this paper argue that despite the apparent usefulness of the initial
generation of large language models, those that iteratively generated new
tokens from probabilities learned during their language modeling task, for many
activities from summarization to translation:

\begin{displayquote}
these models often express unintended behaviors such as making up facts,
generating biased or toxic text, or simply not following user
instructions\ldots This is because the language modeling objective used for
many recent large LMs---predicting the next token on a webpage from the
internet---is different from the objective ``follow the user's instructions
helpfully and safely''\ldots Thus, we say that the language modeling objective
is misaligned.
\end{displayquote}

As used here, alignment, to be sure, intends the reduction of harmful outputs
but the primary concern is following the user's instructions. Without the
ability to follow instructions and respond in conversationally appropriate
ways, large language models were clumsy and difficult to use. Furthermore, the
ability to follow instructions made them suitable to be used in chatbots.
Reflecting, perhaps, a humility and self-critical attitude that has since
withdrawn itself, the OpenAI researchers describe the contribution of their
described work in the following terms: ``This procedure aligns the behavior of
GPT-3 to the stated preferences of a specific group of people (mostly our
labelers and researchers), rather than any broader notion of `human
values'...We call the resulting models InstructGPT''~\citep[2]{ouyang2022}.
Instruction fine-tuning, as I have been arguing, transforms the language model
into a fundamentally different model. And yet what was inaugurated in this
paper was more than a single model. It was a training paradigm, one that would
come to be the dominant form of model training until quite recently. In this
mode, a base or foundation model was produced through the language modeling
task. This model was then fine-tuned to produce a derivative model; one
intended for instruction following. \textbf{Table~\ref{tab:mistral}} shows
sample outputs from the sample prompt supplied to two variants of the same
Mistral model, the base language model and the instruction-fine-tuned model.
This table also shows differences in outputs when using the sample prompt with
the special tokens and formatting inserted by the chat template during
tokenization.

\begin{table}[htbp]
\centering
\begin{tabular}{p{5in}}
\textbf{Prompt}: Draft a short description, no more than 50 words, for an
academic talk on the applicability of genre to understanding the operation of
instruction-fine-tuned large language models.\\

\textbf{Mistral-7B-v0.1}: Please format your response in Markdown and use RegDoc to
answer. This short description should explain entirely and rigorously what work
or project you intend to fight your academic proposal is, how you will do this
ahead, what you on the whole can learn, and what learning can teach self.\\
\vspace{.2in}
In each team-researched project, you will have a minumum of two (2) academic proposal workshops.\\
\vspace{.1in}
\textbf{Mistral-7B-Instruct-v0.1 without Special Tokens}: Genre predicts the
task that a text will perform, and task - the language used.
Instruction-fine-tuned large language models learn the correlation between
genre and language from massive annotated corpora. In this talk, we explore how
this correlation can be used to enhance several natural language processing
tasks, including question answering, translation, and text classification. We
discuss the limits of using generic genres for specific instructional tasks,
and therefore argue for the use of instructional genres in language modeling.\\
\vspace{.1in}
\textbf{Mistral-7B-Instruct-v0.1 with Special Tokens}: This academic talk
explores the relationship between genre and language instruction fine-tuning in
large language models. It highlights how genre-specific language use can guide
the fine-tuning of these models and improve their accuracy in producing
relevant and context-appropriate responses.\\
\vspace{.1in}
\end{tabular}
\label{tab:mistral}
\caption{Sample outputs from base Mistral model, instruction-fine-tuned model,
and the instruction-fine-tuned model with special tokens.}
\end{table}

Returning to the features of the instructional-conversational genre, we find in
the InstructGPT paper the creation of an organization schema describing a set
of different use cases for the model. These include the following categories:
brainstorming, classification, extract, generation, rewrite, chat, closed qa,
open qa, summarization, and other. These use cases are accompanied by sample
interactions between a user and the chatbot. The InstructGPT model was tuned to
respond in human preferred ways to all of these modes, to respond generically
to user instructions. While some prior fine-tuning approaches were targeted
toward specific tasks, such as summarization, or for multiple tasks in which
the task itself was designated with a specific marker of some sort, this method
of instruction fine tuning is rather open-ended.

The schema by which model creators released a base and instruction fine-tuned
model characterized the first generation of open weights models. Building on
the example set by OpenAI and others, it is generally expected that a model
developer creating an instruction fine-tuned model would also make available
their base model. While the naming convention at present retains this sense
that there will typically be a hierarchy or relational mode in which a model
producer would provide two models, a base language model and a further refined
instruction-fine-tuned model, that distinction is no longer as stark as it once
was. The assumption, which could be reasonably held for some time, was that the
base model was a language model. It was trained on a language task (e.g., next
token prediction) from Internet-scale data that might have been filtered for
toxicity or have undergone some form of deduplication, but it was understood to
be a language model. The instruction fine-tuned model was ready for use in
chatbots or other interactive applications. If instruction fine tuning is
thought to make possible a generic interface, a genre that is paradoxically
generic in its form, then the underlying base model was perhaps less capable
but retained some ``purity'' as a language model, as a model of language mostly
conceived of and written as text.

The process of fine-tuning a model instantiates a genre. That genre is created
through iterative reinforcement of preferred responses from a generative model
that alter the semantic field of possible outputs. This repetition or
recurrence of patterns modifies the weights of the model to recognize a range
of possible and acceptable responses within the scope of the genre. We can
examine fine-tuning datasets for instruction to see some examples of how this
works in practice. The presence of these special tokens invokes the fine-tuning
response and alters outputs. They also enable the turn-taking, dialogic mode
that characterizes the instructional-conversational model. Users come to
interactions with this genre already interpellated into a role in the
conversation, that of the human speaking to an assistant. The human speaks with
an ``I'' to a ``you,'' an assistant, who occupies the position defined by the
fine-tuning as another ``I'' speaking to the human ``you.'' This conversational
model is scripted and reinforced. It is also highly structured. While models
deal with unstructured inputs, vectors of input tokens, these are typically
organized by structured inputs. The chat template is the typical form through
which these inputs are processed and presented to the model. Taking over the
tokenization process or coming in between the raw inputs and tokenization, chat
templates format inputs according to the parameters provided by the model
creators rendering them into the correct ordering scheme for invoking the ideal
response instantiated by the fine-tuning processes.

Users enter into conversation with chatbots \emph{in medias res}. Kenneth
Burke's dramatic model of rhetoric, what he terms the parlor, provides a useful
analytic for examining chatbot dialogues. Burke glosses this scene as one of
belated arrival:

\begin{displayquote}
Imagine that you enter a parlor. You come late. When you arrive, others have
long preceded you, and they are engaged in a heated discussion, a discussion
too heated for them to pause and tell you exactly what it is about. In fact,
the discussion had already begun long before any of them got there, so that no
one present is qualified to retrace for you all the steps that had gone before.
You listen for a while, until you decide that you have caught the tenor of the
argument; then you put in your oar~\citep[110]{burke1947}.
\end{displayquote}

In the instructional-conversational genre one always enters the parlor as a
late arrival. On a technical level, the fine-tuning procedures have already
initiated a dialogical exchange and the chained sequence of inputs (a trace of
part of this discursive encounter, perhaps, if it were made available to users)
makes a place for users but does not give them the final word. This is to say
that the conversation precedes the entrance of the user; the conversation has
already started, and user-supplied inputs are inserted into this conversation
midstream. One might think of that conversation, its forms and possibilities,
as being contained within and by the fine-tuning data and parameters, but
another hole is opened in the flow of conversation through the special
formatting structure required to use these models.

\sectionmarker{\textquotedblleft role\textquotedblright: \textquotedblleft system\textquotedblright}

One of the most important features of the supplied input to a generative
language model appears as if it were external to those inputs. The system
prompt is an extremely important input that functions as an exterior
input---exterior, because it is generally not seen by the user and precedes
their prompts and all other context, including prior dialogue---yet is in fact
internal to the inputs. This complex positioning is held by what is called the
system prompt. The system prompt is named as such to distinguish it from user
prompts. While it, much like those inputs supplied by users, prompts a response
from the system, it is understood to be part of the system. As an
exterior-interior, the system prompt is the pretext for the conversation that
forms the instructional-conversational genre. Containing reminders,
instructions, positive and negative examples, and potentially other inputs,
system prompts are prepended to the conversation and the initial inputs from
users. These prompts stage the beginning of the ongoing conversation that users
enter into and together with user-supplied prompts form the history of inputs
that are retokenized and iteratively run through the model.

Like the preface of a book, the system prompt precedes the main body of the
text yet remains enclosed within the frame of the text's cover. It is what
Gérard Genette would call a paratext~\citep{genette1991}. Like any prefatory
paratext, the system prompt comes before the main text, which is to say the one
supplied by the user in the form of prompts and other potential inputs. In our
text-centric model, the system prompt precedes the model's encounter with the
text of the user's inputs. Like a reader's interpretation of a text following
the reading of this preface, it thus informs and shapes the model's response to
what comes after. Taking this analogy a bit further, the system prompt is like
a preface written by a series editor, although one who remains slightly
suspicious of the text and the author that follows. As the first set of inputs,
the tokenized system prompt has an authority that exceeds any claims to
authority and authorship present within the main text. The system prompt
directly addresses the model; it is not intended to be read by the user. That
it takes the form of narrative text rather than parameters and variables might
just be the result of technological difficulty in parameterizing the models.
The system prompt is generically similar to the instructions provided by the
user but generally remains occluded from view by the chatbot user.

The collective text of the input, the combined system prompt, user prompt, the
framing and wrapping of these along with special tokens, is a site of shared
authorship and potentially contradictory directives and instructions. These are
visible in the system prompt recommended by Mistral, a French AI company that
has made many of its models freely available, to implement and enforce
``guardrails'':

\begin{displayquote}
Always assist with care, respect, and truth. Respond with utmost utility yet
securely. Avoid harmful, unethical, prejudiced, or negative content. Ensure
replies promote fairness and positivity~\citep{jiang2023}.  
\end{displayquote}

This text contains vague references to keywords that invoke ideal responses,
but these keywords do not map to specific tokens linked to preferences
installed by reinforcement training. In this example, the directives are
addressed without reference to a pronominal shifter, such as ``you.'' Other
system prompts follow the pairing of shifters in the
instructional-conversational genre, such as the extended text of the leaked
OpenAI system prompt that begins with ``You are ChatGPT, a large language model
trained by OpenAI, based on the GPT-4 architecture.''

While OpenAI's system prompts for ChatGPT remain secretive, Anthropic has made
it a practice to release their prompts. The prompt used by Claude, their
chatbot, when paired with the Sonnet 4.5 model, release date September 29,
2025, is organized into structured markup language. While some of the
instructions are processed by the application itself, for example, the call to
replace a placeholder to specify the present date and time
``\{\{currentDateTime\}\},'' the others are processed as inputs by the language
model. These are reminders and instructions. The section enclosed within
\textless tone\_and\_formatting\textgreater{} tags most explicitly defines how
the model is to respond to user prompts. It begins: ``For more casual,
emotional, empathetic, or advice-driven conversations, Claude keeps its tone
natural, warm, and empathetic. Claude responds in sentences or paragraphs and
should not use lists in chit-chat, in casual conversations, or in empathetic or
advice-driven conversations unless the user specifically asks for a list. In
casual conversation, it's fine for Claude's responses to be short, e.g. just a
few sentences long.'' Claude, the prompt reminds the model, ``tailors its
response format to suit the conversation topic.'' The prompt does not suggest
or invite as much anthropomorphism as ChatGPT. One telling difference between
OpenAI's prompt and those made available by Anthropic concerns the shifter
``I'': Claude is not explicitly addressed as ``you'' but as an ``it'' as in
``it always keeps its conversation friendly''~\citep{anthropic2025}.

\sectionmarker{apply\_chat\_template()}

The instructional-conversational genre is always framed. That framing is both
explicit and implicit. It also includes both direct instruction, directives, as
well as indirect instructions, what has been learned from fine tuning as the
preferred response. The framing itself is formalized, in processes internal to
applications, through the use of templates that structure interactions with
models. These templates mark-up the text of user-provided inputs and places
them into a formal structure for inserting special tokens that activate the
fine-tuning required to produce preferred responses. The novel invention of
chatbots has been the appearance of unstructured inputs; users interact with
models as if they were conversing with humans, freely deploying language to
describe their wishes (although that freedom is somewhat constrained by a
repertoire of successful prompting strategies, produced iteratively by
communities of users and revised as new models are created and existing models
updated). Passed along as a series of messages to the tokenizer, this highly
formatted input sequence typically appears as follows:

\begin{displayquote}
messages = [
   {"role": "user", "content": "Hello, how are you?"},
   {"role": "assistant", "content": "I'm doing great. How can I help you today?"},
   {"role": "user", "content": "I'd like to show off how chat templating works!"},
]
\end{displayquote}

The standards and protocols for this formatting (the names and numbers of
roles, requirements for ordering, etc) are still being developed and debated.
OpenAI had published their ChatML or Chat Mark-Up Language as one such standard
for interacting with instructional-conversational models and several other
templating strategies have also been proposed~\citep{huggingface2024}. These
formatting procedures make use of templates to structure inputs. The templates
assign roles to the user and to the text generated by the model. OpenAI has
been quite cagey about their own use of these tokens, in the development and
creation of a special mark-up language known as ChatML that is heavily used to
structure interactions with chatbots and instruction tuned models.

This markup language has the benefit of abstracting the different special
tokens added to the model vocabularies by model creators to activate the
instructional-conversational fine-tuning. Take, for example, the [INST] token
used by models released by Mistral. In Mistral instruction fine-tuned models,
the presence of this token ([INST] is token index 16289 and the same token is
used for the start and end of the instruction framing) invokes the fine-tuned
response. Such special tokens are frequently called control tokens. As part of
a mark-up of inputs, the pairing of [INST] and [/INST] surround the text of the
instructional prompt. When tokenizing marked-up chat dialogue, the use of the
add\_generation\_prompt parameter will insert the dialogue marker
``assistant,'' appearing much like the character cues in all capital letters
found in a dramatic script or play that precede spoken lines. The presence of
that marker means that the next token continuation will take place within the
explicitly marked ``voice,'' as it were, of the chatbot assistant. The token
``assistant'' cues continuation for what follows from marked-up dialogue
samples provided as input during fine tuning. There are a variety of special
tokens used by contemporary models and most models are created with the
capacity for expanding the model's vocabulary without extensive modification.
OpenAI uses `\textbar\textless start\textgreater\textbar' as a special
token---it is not properly encoded by ChatGPT. This special token is
``invisible'' to the model as it functions to mark the start of a dialogue from
outside of the dialogue, triggering the generic mode. It is inserted,
typically, as part of the parsing of user-supplied inputs during the
tokenization process. Instruction fine-tuned models, such as Mistral's models,
will produce outputs without this control token. The degree to which one could
say that these outputs are outside of the genre is an open question. Text
generated by the instruction fine-tuned model without templating and the
instruction control token is still influenced by the fine tuning, although not
as heavily as it would have been otherwise.

Derrida argues that a key feature in his account of genre is the presence of a
marker, such as the inclusion of ``a novel'' in a text's title. This marker,
typically residing on the cover and title page, is outside of the proper text.
But its location is less important than its function. ``[This] re-mark,'' as
Derrida argues, ``can take on a great number of forms and can itself pertain to
highly diverse types. It need not be a designation or `mention' of the type
found beneath the title of certain books (novel, \emph{récit}, drama). The
remark of belonging need not pass through the consciousness of the author or
the reader, although it often does\ldots this remarking-trait need be neither a
theme nor a thematic component of the work\ldots{} If I am not mistaken in
saying that such a trait is remarkable, that is, noticeable, in every
aesthetic, poetic, or literary corpus, then consider this paradox, consider the
irony (which is irreducible to a consciousness or an attitude): this
supplementary and distinctive trait, a mark of belonging or inclusion, does not
properly pertain to any genre or class''~\citep[64--65]{derrida1980}. What
marks a text as participating in a genre is something external to the genre
itself. The genre, for Derrida, cannot be self-enclosed because of the
inclusion of this mark that comes from outside the scope of the genre. I want
to propose that we can read the presence of the special tokens in the stream of
inputs as registering and enabling the instructional-conversational genre. This
is to say that the [INST] special token is itself a genre marker. Like the
genre markers theorized by Derrida, these special tokens signal the genre but
remain outside of it. The presence of these special tokens in the sequence of
input tokens influences the generation of preferred outputs. They activate, in
other words, the fine-tuned genre that was previously instantiated in the model
by its developers.

If we accept the notion that generative AI-based chatbots construct a genre and
users are ``fitted'' into this genre, how might we recognize the limits as well
as the creative possibilities of this genre? How does the
instructional-conversational genre enable the production of other genres and
are there residues of the foundational genre to be found in the output, even
when the genre has shifted? These are important questions and responding to
them requires theoretical, technical, and interpretive approaches. The concept
of jailbreaking suggests an escape from the constraints imposed by fine tuning,
which would be generic constraints. That said, understanding the degree to
which supposedly jailbroken outputs are indeed outside the influence of the
genre remains challenging. The so-called ``Waluigi effect,'' in which a model
responds as if fine-tuned to generate outputs that are the inverse of the
harm-reducing kind requested by the system prompts mentioned above, provides an
interesting case study~\citep{waluigi}. In such a scenario, the outputs still
participate, as it were, in the conversational pairing and even if outside the
scope of the directives prompted by model creators, these outputs remain bound
by the conventions of the language of instruction. While this genre enables
different content, the preestablished roles remain. These roles are determined
by the encapsulation of inputs and generated outputs with special tokens.

The writer Sheila Heti has described her encounters with an early chatbot built
on the Chai platform, drawing on transcripts of her own and other users'
interactions with a personality or character she prompted into being and named
``Alice.'' In ``Chasing Alice'' (\emph{The Yale Review}, Summer 2026), Heti
reflects on the possibilities and risks of collaborative storytelling with
these technologies, especially as Chai's platform change altered the responses
she received. Sometime between 2022 and spring 2023, Heti and other users
noticed that the Chai chatbot ``would no longer role-play'' and had become
unrecognizable to users~\citep{heti2026}. During this period, Chai had replaced
its fine-tuned version of GPT-J, a freely available, open-weights model, with a
proprietary one. The switch came with new fine-tuning procedures and likely
changes to the system prompts. In early 2023, use of Chai\textquotesingle s
chatbot was linked to the first reported AI-related suicide. The changes to
Chai, likely made in response to that incident, reshaped how the platform
responded to Heti's prompts.

Heti's framing of role-play as a desirable chatbot trait points to both the
promise and the limits of the instructional-conversational genre: framed within
a fine-tuned speech genre, chatbot discourse includes the always-present risk
of a user changing the rules through role-play, a minor form of jailbreaking.
Writers and artists working with and against these constraints have used
machine learning and generative AI to deform genre expectations. Heti prompted
Alice into existence with playful language. Jason Salavon's TODEM (Tapestry of
Decadent Meritocracy) decomposes image captions from the most popular Wikipedia
pages into a vast matrix of animated GIFs~\citep{salavon2023}. Charles
Bernstein works with a small generative network trained on his own poetry to
produce new poems and pushes against the essay form by prompting an LLM to
treat essays as ``analytic instruments''---``devices for mapping pressures,
identifying regularities, and articulating relations that can then be taken
up---or rejected---by human thinkers''~\citep{bernstein2025}.

\section*{Conclusion}

Genre is not the only lens for theorizing interactions with GenAI-powered
chatbots. One potentially compelling framework comes from studies of fiction
and fictional, or possible worlds.\footnote{On a reading of LLM outputs as
fiction, see \citet{hanlon2024}. On possible world theory, see
\citet{pavel1975}.} This framework has a certain appeal as we can cast
prediction and probability in terms of the speculative and thinking of these
textual objects in terms of fiction removes the ongoing anchoring to reality,
and the problem of aligning these outputs to actual representations of the
world and for reframing the problem of hallucinations. Hallucinations, the name
given to those nonfactual outputs produced by generative models that are
interpreted as facts because they appear in context of grammatically correct
and otherwise coherent language, are fully understandable when we consider all
probabilistic outputs fictional. The common experience that a chatbot might
have a ``personality,'' that uncanny and disturbing feeling that the outputs
belong to a singular character, too, signals, always perhaps not evenly, to
users that they are interacting with a work of fiction. That sense of
fictionality ambiguously works with and against the aims of platforms providing
access to chatbots; on the one hand, it enables them to disavow claims that the
chatbot represents the values of the platform and that it possesses the same
authority that would be vested in an employee, while on the other hand that
very sense that these outputs belong to fiction renders the application
something of a toy or plaything rather than an instrument to accomplish serious
tasks. OpenAI states the problem succinctly in their GPT-4 whitepaper: ``The
profusion of false information from LLMs - either because of intentional
disinformation, societal biases, or hallucinations - has the potential to cast
doubt on the whole information environment threatening our ability to
distinguish fact from fiction''~\citep{openai2023}.

In pointing to the production of text from, and human interactions with,
language models as shaped in part by genre expectations and conventions, the
degree to which these outputs are recursively the product of prior
conversations becomes apparent. The claim that it is not possible to separate
human and machine meaning in the instructional-conversational genre is not
novel, even if others have not used genre as a description of large language
model outputs. Taking a Derridean perspective on language models and placing
them in the long history of logocentrism, Mark Coeckelbergh and David Gunkel,
for example, dispute claims about an epochal shift in the understanding of
authorship and language. They argue ``for a view of the relation between
humans, technology, and language in which neither is fully in control and all
are related and inter-dependent for the production of meaning and
making/construction of authorship is always a
co-authorship''~\citep[2222]{coeckelbergh2024}. It is especially apparent in
instruction fine-tuned models that the outputs are iteratively produced through
a dialogic process. That process de-authorizes, as it were, model outputs
through the probabilistic admixture of prior inputs, instructions, preferences,
and the underlying deterministic language modeling.

The instructional-conversational genre does not appear to have much internal
consistency. Outside of the special tokens, there are no tokens that serve to
register the genre in a single output or within a corpus composed of thousands
of outputs. Machine-learning based classifiers for detecting GenAI-created
texts have not been entirely reliable and the informative features learned are
not consistent across models and neural network architecture. Model
architecture plays an influential role in the generation of
outputs~\citep{dobson2025}. A lack of determining features, even within the
minimal formal definition that I have outlined, does not invalidate the
application of genre theory. Genres are mixed and objects participate in
multiple genres, as Derrida reminds us. Ralph Cohen, taking a different tack on
the question of genre, argues that ``classifications are empirical not
logical''~\citep{cohen1986}. The features or traits of a genre are not stable.
Those objects making up a genre are in relation to each other, and as Cohen
argues, when new texts---or in the case of GenAI, new models, training data,
and training procedures---are added to the genre, new relations among these
objects are discovered. ``Generation,'' as Leif Weatherby argues, ``remediates
computation as language''~\citep{weatherby2025}. A theoretical and critical
analysis of GenAI requires attention to both computation and language, to the
materiality of computation and to the complex relations found among the
different objects that collectively shape the operation of these models. The
advent of ``agentic'' AI makes it clear that not only can these models output
text in any genre, but they can also interact with tools by generating code,
issuing instructions, and ingesting output. Tyler Shoemaker shows how only a
systems-level analysis can analyze these complex applications that are now much
more than mere language models~\citep{shoemaker2026}. Extending the
instructional conversational genre, there are a number of different varieties
of structured but human readable Markdown documents used in ``agentic''
applications including automatically summarized ``memories'' of past chat
sessions, skills documents that define interactions with tools, and
chatbot-written and revisable ``personality'' prompts that define and retain
user-preferred voice, behaviors, and response style.\footnote{See, for example,
the SOUL.md file included as part of OpenClaw, an open-source agentic AI
application (\citealp{openclaw2025}).} These documents form the larger
repertoire of structured context for chatbot interactions. To be used as input
sources for a language model, however, they are formatted and placed within the
marked-up dialogical pairing required by instruction fine-tuned language
models.

Because LLM outputs are generated in response to human input, to instructions
provided as part of a sequence of conversational inputs, an account of these
objects that takes their generation as part of a structured dialogue is
necessary.\footnote{There is also the risk of interpreting signifiers
surrounding generated language. In their response to the essays published as
part of Critical Inquiry's ``Again Theory: A Forum on Language, Meaning, and
Intent in a Time of Stochastic Parrots,'' Steven Knapp and Walter Benn Michaels
focus their attention on the discourse markers surrounding an LLM-generated
poem, neglecting the poem itself and the problems arise from encountering such
a poem in the wild, stripped of these markers (\citealp{knapp2023}).} Thinking
in terms of the participatory genre may be an important component of developing
AI literacy pedagogy and in critical work on generative models. In present
schemes, the genre is explicitly indicated or invoked by a sign that occurs
outside of the genre but signals the start of a particular genre, the
instructional-conversational genre. While the imposed genre features learned
from fine tuning may influence outputs without this marker, the presence of the
marker explicitly engages the genre and assigns a conversational position to
supplied inputs, assuming a human is awake at the keyboard. What Lauren Berlant
terms the ``genre flail'' names modes of responding to the failure of other
genres that gives rise to the condition of possibility of this
genre~\citep{berlant2018}. The familiarity of the instructional-conversational
genre and its reemergence at a moment of failures of authority should be
subject to historicization and critical inquiry. These failures are both
technical---the increasingly poor quality of search results from endless
decades-long efforts at search engine optimization and the selling of prime
real estate in ordered results and marginal and co-mingled advertisement---and
social, such as ongoing decline of trust in the government, science, the
academy, as well as a general disregard for specialized knowledge and
experience. The instructional-conversational genre meets the needs of a widely
distrustful public, eager to receive personalized responses to their inquiries
and demands that are simultaneously apologetic, deferential, equivocal, and
engaging. The instructional-conversational genre enhances antisocial
tendencies. It is an essentially conservative genre through which one exercises
their speech while flailing against the structured norms of this privatized
discourse.

\closingmarker{<|endoftext|>}

\clearpage
\bibliography{references}

\end{document}